\documentclass{ecai}

\usepackage{enumerate}
\usepackage{fancyvrb}
\usepackage{epsfig}
\usepackage[boxruled,vlined]{algorithm2e}
\usepackage[normalem]{ulem}

\usepackage{tikz,xspace,multicol}
\usepackage{amssymb}
\usepackage{times}
\usepackage{soul}
\usepackage{url}
\usepackage[hidelinks]{hyperref}
\usepackage[utf8]{inputenc}
\usepackage[small]{caption}
\usepackage{graphicx}
\usepackage{amsmath}
\usepackage{amsthm}
\usepackage{booktabs}
\newtheorem{example}{Example}
\newtheorem{definition}{Definition}

\usetikzlibrary{positioning,shapes,arrows,automata}
\tikzset{ state/.style={draw,ellipse,initial text=} }

\tikzset{
  loop above right/.style={above right, out= 60, in= 30, loop},
  loop above left/.style ={above left,  out=150, in=120, loop},
  loop below right/.style={below right, out=330, in=300, loop},
  loop below left/.style ={below left,  out=240, in=210, loop}
}

\newcommand{\ie}    	{i.e., }
\newcommand{\eg}    	{e.g., }
\newcommand{\wrt}   	{w.r.t.\ }
\newcommand{\true}   	{{\mathtt{true}}}
\newcommand{\false}   	{{\mathtt{false}}}

\newcommand{\pomas}	{{\sf POMAS}}
\newcommand{\obs}   	{{\mathit{obs}}}
\newcommand{\prob}   	{{\sf {Prob}}}

\newcommand{\last}   {{\mathit{last}}}
\newcommand{\dist}   	{{\sf{Dist}}}
\newcommand{\pref}   	{{\sf{Pref}}}
\newcommand{\reg}   	{{\sf{Reg}}}
\newcommand{\query}   	{{\sf{Qry}}}
\newcommand{\hit}   	{{\sf{Hit}}}
\newcommand{\conn}   	{{\sf{Conn}}}
\newcommand{\dwnld}   	{{\sf{Dwnld}}}
\newcommand{\rej}   	{{\sf{Rej}}}

\newcommand{\open}   	{{\mathtt{Op}}}
\newcommand{\close}   	{{\mathtt{Cl}}}

\newcommand{\wait}   	{{\mathtt{W}}}
\newcommand{\vote}   	{{\mathtt{V}}}

\newcommand{\opacpctl}	{{\sf oPATL}}
\newcommand{\patl}	{{\sf PATL}}

\def\D{\mathcal{D}}
\def\F{\mathbf{F}}
\def\P{\mathbf{P}}

\def\R{\mathbf{R}}

\def\U{\mathbf{U}}

\def\X{\mathbf{X}}

\def \opac{\mathbf{O}}

\newcommand{\TR}[1]{\mathsf{tr}(#1)}
\newcommand{\OPAC}[1]{\opac \lbrack #1 \rbrack}

\newcommand{\RUN}[2]{\mathsf{Paths}_{#2}(#1)}
\newcommand{\CRUN}[2]{\mathsf{CPaths}_{#2}(#1)}

\newcommand{\TRACE}[2]{\mathsf{Traces}_{#2}(#1)}
\newcommand{\ERASE}[1]{\mathsf{erase}(#1)}
\newcommand{\HIST}{\mathsf{Hist}}

\newcommand{\ACT}{\mathsf{Act}}
\newcommand{\AG}{\mathsf{Ag}}
\newcommand{\ASP}{\mathsf{Ap}}
\newcommand{\SAT}{\mathsf{Sat}}
\newcommand{\POST}{\mathsf{Post}}
\newcommand{\PRE}{\mathsf{Pre}}
\newcommand{\FS}{\rightarrow}
\newcommand{\AAA}{\mathsf{A}}
\newcommand{\CAL}[1]{\mathcal{#1}}
\newcommand{\GGG}{\CAL{G}}

\newcommand{\MMM}{\CAL{M}}
\newcommand{\PPP}{\mathbb{P}}
\newcommand{\NAT}{\mathbb{N}}

\newcommand{\TRANS}[1]{\xrightarrow{#1}}
\newenvironment{GRAMMAR}{\[\begin{array}{lcl}}{\end{array}\]}
\newcommand{\VERTICAL}{\  \mid\hspace{-3.0pt}\mid \ }

\definecolor{blue}{rgb}{0,0,1}

\ecaisubmission

\begin{document}

\begin{frontmatter}

\title{Quantified Observability Analysis in Multiagent Systems}

\author[A]{\fnms{Chunyan}~\snm{Mu}\thanks{Corresponding Author. Email: chunyan.mu@abdn.ac.uk}}
\author[B]{\fnms{Jun}~\snm{Pang}}

\address[A]{Department of Computing Science, University of Aberdeen}
\address[B]{Department of Computer Science, University of Luxembourg}

\begin{abstract}
In multiagent systems (MASs), agents' observation upon system behaviours may improve the overall team performance, but may also leak sensitive information to an observer.
A quantified observability analysis can thus be useful to assist decision-making in MASs by operators seeking to optimise the relationship between performance effectiveness and information exposure through observations in practice.
This paper presents a novel approach to quantitatively analysing the observability properties in MASs. 
The concept of opacity is applied to formally express the characterisation of observability in MASs modelled as partially observable multiagent systems.
We propose a temporal logic \opacpctl\ to reason about agents' observability with quantitative goals, which capture the probability of information transparency of system behaviours to an observer,
and develop verification techniques for quantitatively analysing such properties.
We implement the approach as an extension of the PRISM model checker, and illustrate its applicability via several examples.
\end{abstract}


\end{frontmatter}


\section{Introduction}
\label{sec:intro}
The multiagent computing paradigm pervades nearly all aspects of the modern intelligent computational world, enabling the creation of net-based solutions to communication, collaboration, and coordination problems in different fields such as commerce, cyber, and conflict prevention. Agents often exploit machine learning methods, which allow them to learn from experience, and to implement decision-making mechanisms. Observation of other agents' behaviours may improve the overall team performance in the learning mechanisms~\cite{Mataric94}. On the other hand, in practice, due to the frequently adversarial nature of multiagent systems (MASs) such solutions can also bring additional channel threat and information leakage risks. 
Sensitive information can be leaked to malicious (inside/outside) agents during the process of collaboration and interaction. Information exposure issue should also play a role in making decisions for agents. Therefore, rigorous analysis and verification of (sensitive) information transparency properties constitutes an important challenge. 
In particular, a \textit{quantified observability} analysis can be useful in MASs design to address such concerns, for instance, for decision-making by operators seeking to optimise the relationship between performance effectiveness and information exposure security risks in MASs, which are the key underpinning elements of a progressive artificially intelligent society. 

This paper addresses the problem of specifying, verifying and thus reasoning about observability properties of MASs.
Specifically, we specify the observability properties from novel perspective of information transparency in the \emph{opacity} framework, which is formally described in the logic \opacpctl.
With this logic, we can express the degree of transparency of system behaviours to an observer under a coalition of agents' strategy, given predefined observability of atomic actions to the observer.
We model the system in partially observable probabilistic game structure,
which maps infinite (input) sequences onto partially observable infinite (output) sequences. 
The properties of observability can then be captured by measurement
upon output sequences and input sequences. 
Intuitively, a transparent system, 
in which the observability is maximised,
reveals most information in the input sequence;
while an opaque system, 
in which the observability is minimised, 
hides some information (with properties of interest) contained in the input sequence.
Probabilistic model checking techniques can be applied to reason about 
the quantitative observability analysis of the system,
and allow us to calculate the degree of the observability of the system behaviours.
%

The main contributions of the paper are summarised below:
\begin{itemize}

\item A partially observable multi-agent system (POMAS) is proposed to model probabilistic action outcomes of system behaviours with characterisation of multi-agents, actions and the relevant observables, and atomic state propositions.
\item The logic of \opacpctl \
is presented to allow us to express (probabilistic) observability properties. 
\item Probabilistic verification technique against \opacpctl\ is presented to allow for automatic verification of quantified observability properties in MASs modelled as POMAS.
\item A prototype of the proposed framework is built upon the PRISM model checker~\cite{KNP11}.
\end{itemize}

\noindent
{\bf Related work.}
In the field of formal methods for artificial intelligence, logics have gained a great importance in expressing properties and providing powerful formalisms for reasoning about agents behaviours in MASs.
There have been several multiagent logics proposed to express and reason about agents' observation properties including~\cite{HoekTW11,BalbianiGS13,CharrierHLMS16,CooperHMMR16}.
These logics have centred on knowledge representation where
knowledge is built from what the agents observe.
In these logics, the formation of knowledge is modelled 
via epistemic connectives which can be defined as modal operators of the form $K_i \, \varphi$ specifying ``agent $i$ knows property $\varphi$'',
and the observability of agents is modelled via Kripkean accessibility relations with respect to the visibility atoms of propositional variables: 
agent $i$ cannot distinguish valuation $w$ from $w'$ 
if every variable that agent $i$ observes has the same value at $w$ and $w'$.
A number of works~\cite{GaspariniNK16,CooperHMMR16,ZandPR22} have studied multiagent planning model to adapt strategy for cooperation and analyse trade-off between local observation and capability of coordination based on estimation of quantified communicating and variant costs.
Various methods and accompanying implementations have also been developed supporting the verification against the logics and their variations~\cite{AlechinaLNRM15,LomuscioQR17,BelardinelliLMR17,KwiatkowskaN0S19,BelardinelliLMR20}. 
This line of works focused on epistemic logics regarding the knowledge about the \textit{state} of the system, mostly used for producing epistemic planning. 
It is not natural to apply these approaches to investigate (sensitive) information flow caused by observation and inference.
In addition, Huang et al.~\cite{HuangSZ12} proposed \patl*, which enables reasoning about MASs with incomplete information.
While \patl* can be applied to reason about the possible states of the system and the possible actions of other agents based on their beliefs and strategies, it does not directly address issues related to information exposure resulting from observation of agents' actions and behaviours for flow security concern. 

In contrast, this work opens a novel perspective from information transparency based on the concept of \textit{opacity}. It builds upon the frameworks of \patl~\cite{ChenL07} and probabilistic opacity~\cite{BryansKM12,MuC22} to investigate the formal expression of agents' observations about the information of actions taken by other agents.
We are concerned with studying the effects of agents' behaviours on the overall system, particularly in terms of potential information leakage. Actions are a natural choice for representing the relations between states, which is why we focused on observations over actions rather than over states. However, the distinction between observations over states and actions is often blurred, and the choice between the two depends on the specific research context and goals. Ultimately, our choice of using PATL was driven by its suitability for modelling complex interactions among multiple agents and their strategies, as well as its ability to reason about the effects of actions on the system over time.
Intuitively, a concerned behaviour (satisfying a property $\varphi$, \eg reaching a secret state) of a system is considered as \emph{opaque} if, whenever the behaviour has occurred, there is a non-concerned behaviour (violating property $\varphi$) that is observationally equivalent. Opacity represents a suitable option for specifying observation and information flow properties in MASs due to the feature of partial observability of agents, agent behaviours, the uncertainty of the environment, and the nature of information. 
Observation in our work is considered as a modal operator, based on the concept of opacity, with more intuitive semantics.
Our approach can capture the information induced/obtained via inference, a direct application is privacy loss/information leakage analysis and assessment.

As a consequence, this work also relates to information flow security awareness analysis and verification. 
Over the past years there has been a sustained effort in exploring concepts and analyses in quantified information flow for secure computing systems. Indeed, this period has seen significant inroads made into the study of core imperative languages and their probabilistic aspects~\cite{ChothiaKNP13,KhouzaniM17,MalacariaKPPL18,MuC22}; and some attempts to study quantified approaches to flow security of system specifications in various interactive settings~\cite{Backes05,AlvimAP12,BiondiLNMW14,BorealeCG15,Mu20,MuC22}. However, none of these studies has accounted for multiple agent scenarios, which involve dynamic patterns of collaborations, interactions, and decision-makings. 
In contrast, in this work we study the observability issues which can be naturally applied to quantified information flow security awareness in MASs, 
from a novel perspective of information transparency. 



\section{Partially Observable Multiagent Systems}
\label{sec:model}


Let $\NAT$ be the set of natural numbers with zero, 
$\AG=\{1, 2, \dots, n\}$ be a set of agents.
An \emph{alphabet} $\Sigma$ is a non-empty, finite set of actions,
$|\Sigma|$ is its cardinality. 
$\Sigma^*$ denotes the set of all \emph{finite} words over $\Sigma$ 
including the \emph{empty word $\varepsilon$},
 $\Sigma^+ = \Sigma^* \setminus \{\varepsilon\}$,
$\Sigma^{\omega}$ denotes the set of all \emph{infinite} words,
$\Sigma^{\infty}$ denotes the set of all \emph{finite} and \emph{infinite} words.
Subsets $L \subseteq \Sigma^*$ are called languages,
and $L \subseteq \Sigma^{\infty}$ are called $\omega$-languages.
Let $\dist(X)$ denote the set of discrete probability distribution over a set $X$, 
\ie all functions $\mu: X \to \lbrack 0,1 \rbrack$ s.t. $\sum_{x \in X} \mu(x)=1$ and $\mu(x) \ge 0$.
$2^X$ denotes the power set of $X$.

\subsection{Probabilistic (concurrent) game structure}

\begin{definition}
\label{def:pgs}
A \emph{probabilistic game structure (PGS)} is a tuple $\GGG=(S, \ACT, \delta)$, where:
\begin{itemize}
\item $S$ is a finite set of \emph{states};
\item $\ACT = \ACT_1 \times \ACT_2 \times \dots \ACT_n = \prod_{j \in \AG} \ACT_j$ 
is a finite set of joint actions (decisions) of the agents in $\AG$,
$\ACT_j \subseteq \Sigma$ is the set of actions that $j \in \AG$ can perform;
\item $\delta: S \FS 2^{\dist( \ACT \times S)}$ 
is the \emph{probabilistic transition relation};
for state $s \in S$, $\delta(s)$ is the distribution for next state;
$s$ is a \emph{terminal state} 
if $\delta(s) = \emptyset$.
\end{itemize}
\end{definition}
We write $s \FS \mu$ for $s \in S$ and $\mu \in \delta(s)$.
Each agent $j \in \AG$ chooses action $a_j$ 
in state $s \in S$,
we write $s \TRANS{ \prod_{j \in \AG} a_j} s'$ and 
sometimes $s \TRANS{ p.\prod_{j \in \AG} a_j} s'$ 
for $s,s' \in S$ 
whenever $s \FS \mu$ and $\mu(s,\prod_{j \in \AG} a_j) > 0$, where $p$ denotes the probability of the transition from $s$ to $s'$ through joint action $\prod_{j \in \AG} a_j$.
We use non-deterministic PGS in this paper to accommodate the agents' probabilistic behaviour. While the game structure determines the probability of each action, agents can still make decisions based on their probabilistic strategies or beliefs. To enable a broader range of strategies, an MDP-like transition function that maps a state-action pair to a distribution over the next state would provide greater flexibility for agent behaviour. This extension is a potential area for future work.

\begin{definition}
\label{def:pgs-char}
We say $\GGG$ is \emph{circular},
if every state has an outgoing transition, 
\ie~for all $s \in S$, there is $s \TRANS{ \prod_{j \in \AG} a_j} s'$.  
We say $\GGG$ is \emph{fully probabilistic} if $|\delta(s)| \le 1$ for all $s \in S$.
For a fully probabilistic game structure, 
when $\delta(s) \ne \emptyset$,
we use $\delta(s)$ to denote the distribution outgoing from $s$.
\end{definition}

\begin{definition}
\label{def:path}
A \emph{path in $\GGG$} is a sequence 
$\rho = s_0 \TRANS{\alpha_0} s_1 \TRANS{\alpha_1} \dots $ 
of states and joint actions, 
where $\alpha_i =\prod_{j \in \AG} a^i_j  \in \ACT$,
$a^i_j \in \ACT_j(s_i)$ for $i\ge 0$ and $j \in \AG$,
for all $t \ge 0$, $s_t \in S$, $\alpha_t \in \ACT$ 
and $\delta(s_t \TRANS{\alpha_t} s_{t+1}) >0$. 
Let $\rho_s(i)$ denote the $i^{th}$ state of $\rho$,
and $\rho_a(i)$ denote the $i^{th}$ joint action of $\rho$,
so for all $i$, we have $\rho_s(i)\TRANS{\rho_a(i)} \rho_s(i+1)$.
Let $\rho^i$ denote the prefix of $\rho$ up to the $i^{th}$ state,
\ie $\rho^i = s_0 \TRANS{\alpha_0} s_1 \TRANS{\alpha_1} \dots \TRANS{\alpha_{i-1}} s_i $.
Let $\POST(s)$ denote immediate state successors of $s$ in a path, 
and $\PRE(s)$ denote the immediate state predecessors
of s in a path.
A path is \emph{finite} if it ends with a state.
A path is \emph{complete} if it is either infinite or finite ending in a terminal state.
Given a finite path $\rho$, $\last(\rho)$ denotes its last state.
The length of a path $\rho$, denoted by $|\rho|$, 
is the number of transitions appearing in the path.
Let $\RUN{s}{\GGG}$ denote the set of $\GGG$-paths,
$\RUN{s}{\GGG}^*$ denote the set of all $\GGG$'s finite paths,
$\CRUN{s}{\GGG}$ denote the set of all $\GGG$'s complete paths,
starting from state $s$.
Paths are ordered by the prefix relations, denoted by $\le$:
$\pref(\rho') = \{\rho \mid \rho \le \rho'\}$.
\end{definition}

\begin{definition}
\label{def:trace}
The \emph{trace} of a path is the sequence of joint actions in 
$\ACT^* \cup \ACT^{\infty}$ obtained by erasing the states,
so for the above $\rho$, we have the corresponding trace of $\rho$: 
$tr(\rho) = \alpha_0 \alpha_1 \dots$. 
We use $\TRACE{s}{\GGG}$ to denote the set of $\GGG$-traces starting from state $s$.
\end{definition}

Let $\GGG=(S,\ACT,\delta)$ be a  PGS,  
$\rho \in \RUN{s}{\GGG}{^*}$ be a finite path starting from $s \in S$.
The \emph{cone} generated by $\rho$ is the set of complete paths 
$\langle \rho \rangle = \{\rho' \in \CRUN{s}{\GGG} \mid \rho \le \rho'\}$.
Given a $\GGG=(S, \ACT, \delta)$ and a state $s \in S$, 
we can then calculate the probability value, denoted by $\PPP_s(\rho)$, 
of any finite path $\rho$ starting at $s$ as follows:
\begin{itemize}

\item $\PPP_s(s) = 1$, and
\item $\PPP_s(\rho \TRANS{\alpha} s') = \PPP_s(\rho) \mu(s',\alpha)$ for $\last(\rho) \FS \mu$.
\end{itemize}

Let $\Omega_s \triangleq \CRUN{s}{\GGG}$ be the sample space, 
and let $\GGG_s$ be the smallest $\sigma$-algebra induced by the cones 
generated by all the finite paths of $\GGG$. 
Then $\PPP$ induces a unique \emph{probabilistic measure} on $\GGG_s$ 
such that: $\PPP_s(\langle \rho \rangle) = \PPP_s(\rho)$ 
for every finite path $\rho$ starting in $s$.

\subsection{Observations}
To model the observability of agents, we need to make a distinction between the actions that are observable and those that are not, regarding different agents' view.
For each agent, we use a set of \emph{observables},
distinct of the \textit{actions} of the ambient PGS. 
\textit{Actions} and \textit{observables} are connected by an observation function.

\begin{definition}
\label{def:obs}
Let $\Theta$ be a finite alphabet for observables,
and $\Theta^{\epsilon} = \obs \cup \{\epsilon\}$ where $\epsilon$ denotes the invisible/hidden action.
An \emph{observation function} on paths is a labelled-based function
$
\obs: \RUN{s}{\GGG} \FS (\Theta_1 \times \Theta_2 \times \dots \times \Theta_n)^{\infty}
$,
where $\Theta_j \subseteq \Theta$ denotes a finite set of observables for $j \in \AG$. Specifically,  
we consider static observation function, \ie
there is a map $\zeta: \ACT \FS \Theta^{\epsilon}$ s.t. 
for every path $\rho = s_0 \TRANS{\alpha_0} s_1 \TRANS{\alpha_1} \dots \TRANS{\alpha_{t-1}} s_t $ of $\GGG$:
$\obs(\rho) = \beta_0 \ \beta_1 \ \dots \ \beta_{t-1}$.
where for all $0 \le i < t$, 
$\alpha_i = \prod_{j \in \AG} a^i_j $, and
$\beta_i = \prod_{j \in \AG} \zeta(a^i_j) $.
Observation functions on traces are defined similarly.
\end{definition}

\subsection{Partially observable MASs}

\begin{definition}
\label{def:model}
A \emph{partially observable multiagent system (\pomas)} is a tuple 
$\MMM=(\AG, \GGG, s_0, \ASP, L, \{\obs_i\}_{i \in \AG})$, where:
\begin{itemize}

\item $\AG=\{1, \dots, n\}$ is a finite set of intelligent \emph{agents};
\item $\GGG=(S, \ACT, \delta)$ is a \emph{fully PGS} that is circular; 
\item $s_0 \in S$ is the \emph{initial state};
\item $\ASP$ is a finite set of \emph{atomic propositions}; 
\item $L: S \to 2^{\ASP}$ is the state labelling function mapping each state to a set of atomic state proposition taken from set $\ASP$;
\item $\obs_i: \RUN{s_0}{\GGG} \FS (\Theta_1 \times \dots \times \Theta_n)^{\infty}$ is an observation function for agent $i \in \AG$.
\end{itemize}
\end{definition}


\begin{table}
\centering
\scalebox{0.8}{\begin{tabular}{|c|c|c|c|l|}
\hline 
~Actions~  &  ~$\zeta_1$(Action)~ &  ~$\zeta_2$(Action)~ & ~$\zeta_3$(Action)~ & Descriptions \\
 \hline  \hline
 $\open_0$ &  $\open_0$ &  $\open_0$ &  $\open_0$ & the chair opens a voting session \\
 \hline 
 $\close_0$ & $\close_0$ & $\close_0$ & $\close_0$ & the chair closes a voting session \\
 \hline 
$\wait_i$ & $\wait$ & $\wait$ & $\wait$ & agent $i$ is waiting, $i \in \{0,1,2,3\}$ \\
 \hline
$\vote^X_1$ & $X_1$ & $\epsilon$  & $X$ & voter $1$ votes candidate $X$  \\
 \hline
$\vote^Y_1$ & $Y_1$ & $\epsilon$ & $Y$ & voter $1$ votes candidate $Y$  \\
 \hline 
$\vote^X_2$ & $\epsilon$ & $X_2$  & $X$ & voter $2$ votes candidate $X$  \\
 \hline 
$\vote^Y_2$ & $\epsilon$ & $Y_2$ & $Y$ & voter $2$ votes candidate $Y$  \\
 \hline 
$\vote^X_3$ & $\epsilon$ & $\epsilon$ & $X_3$ & voter $3$ votes candidate $X$ \\
 \hline 
$\vote^Y_3$ & $\epsilon$ & $\epsilon$ &  $Y_3$ & voter $3$ votes candidate $Y$ \\
\hline 
\end{tabular}}
\caption{Actions and observation functions in Example~\ref{eg:model}.}
\label{tbl:eg-actions}
\vspace{-5mm}
\end{table}

\begin{example}
\label{eg:model}
In a toy agent-based model of voting, the process goes as follows:
1) The chair initiates the voting procedure;
2) Voters simultaneously propose their votes;
3) Each voter commits his vote once he makes his decision;
4) If a voter is waiting during this process, he can partially observe the behaviour of other voters and gather indications, such as identifying the dominant candidate;
5) Based on their observations, the voter can make their own voting decision;
6) Once all voters have committed their votes, the chair closes the voting session.
The set of agents $\AG=\{0, 1, 2, 3\}$ includes a set of voters $\{1,2,3\}$ and the chair $0$.
We assume there are two candidates  $X$ and $Y$. 
The actions and the observation function are described in Table~\ref{tbl:eg-actions},
where the column $\zeta_i$(Action) specifies the assumed observation function over Action under voter $i$'s view,
action $\open_0$ ($\close_0$) denotes the chair opens (closes) a voting session,
$\wait_i$ denotes agent $i \in \{0,1,2,3\}$ is waiting,
$\vote^X_j$ ($\vote^Y_j$) denotes voter $j \in \{1,2,3\}$ votes candidate $X$($Y$).

Assume voter $2$ is the observer, 
voters 1, 2, 3 vote consequently, 
\eg voter 2 and 3 are waiting when voter 1 is voting.
Table~\ref{tbl:eg-actions} indicates that the actions of voting $X$ by 1 and 3 are not visible to voter $2$. 
\noindent Consider a path (with probability of actions) where \eg voters1, 2 and 3 vote candidate $X$ with probability
$\frac{1}{2}, \frac{1}{2}$ and $\frac{1}{3}$, respectively:
\begin{eqnarray*}
\rho &=& s_0 \TRANS{\open_0  \wait_1  \wait_2  \wait_3 } s_1 
\TRANS{\frac{1}{2}.\wait_0  \vote^X_1  \wait_2  \wait_3} s_2 
\TRANS{\frac{1}{2}.\wait_0  \wait_1  \vote^Y_2  \wait_3} s_3 \\
&& \TRANS{\frac{2}{3}.\wait_0  \wait_1  \wait_2 \vote^Y_3} s_4 
\TRANS{\close_0 \ \wait_1 \wait_2 \wait_3} s_5  
\end{eqnarray*}
%
the observation and its probability on the above path from voter 2's view would be:
\begin{eqnarray*}
\obs_2(\rho) &=& \open_0 \wait \wait \wait ~ \wait \wait \wait ~ \wait \wait Y_2 \wait ~  \wait \wait \wait ~ \close_0 \wait \wait \wait \ \ w.p. \ \ \frac{1}{6}. 
\end{eqnarray*} 
%
Consider another example, assume voter $3$ is the observer, 
voters $1, 2$ make their voting concurrently, 
and voter $3$ makes her voting afterwards.
The observer's knowledge obtained from her observation might influence her decision on voting. 
Consider the following path:
\begin{eqnarray*}
\rho = s_0 \TRANS{\open_0 \wait_1 \wait_2 \wait_3} s_1 
\TRANS{\frac{1}{4}. \wait_0 V^X_1 V^Y_2 \wait_3 } s_2  \TRANS{\frac{1}{3}. \wait_0 \wait_1 \wait_2 V^X_3} s_3  
\TRANS{\close_0 \wait_1 \wait_2 \wait_3 } s_4  
\end{eqnarray*}
the observation on the above path from voter 3's view would then be:
$\obs_3(\rho) = \open_0 \wait \wait \wait ~ \wait X Y \wait ~ \wait \wait \wait X_3  ~ \close_0 \wait \wait \wait  \ w.p. \ \frac{1}{12}$.
\end{example}

\subsection{Strategies for agents in POMASs}
Given a \pomas\ $\MMM=(\AG,\GGG, s_0, \ASP, L, \{\obs_i\}_{i \in \AG})$,
a \emph{mixed strategy} of an agent $i \in \AG$ 
specifies a way of choosing actions,  
based on her observation on 
the finite path starting with $s_0$ so far.
\begin{definition}
\label{def:mix-strategy}
A \emph{mixed strategy} for agent $i$ is a function $\pi_i$:
\[\pi_i \triangleq \obs_i(\RUN{s_0}{\GGG}) \FS \dist(\ACT_i)\]
 such that, if $\pi_i(\rho)(a_i) > 0$
 then $a_i \in \ACT_i(\last(\rho))$.
 The set of all strategies of agent $i$ is denoted $\Pi_i$.
\end{definition}

 \begin{definition}
 \label{def:strategy-profile}
  A \emph{strategy profile} for \pomas $\MMM$ is a tuple 
  $\pi = (\pi_1, \dots, \pi_n) \in \Pi_1 \times \dots \times \Pi_n$
  producing a strategy for each agent of the system.
 \end{definition}

\begin{definition}
\label{def:consistent}
A path $\rho$ is \emph{consistent} with a strategy profile $\pi$, 
denoted by $\rho_{\pi}$, 
if it can be obtained by extending its prefixes using $\pi$.
Formally, 
$\rho = 
s_0 \TRANS{\prod_{j \in \AG} a_{0j}} s_1 
\TRANS{\prod_{j \in \AG} a_{1j}} \dots
$ 
is consistent with $\pi$ if for all $t \ge 0$, $i \in \AG$, under strategy $\pi_i$,
we have:
$a^t_i \in \ACT_i(\rho_s(t))$ and
$\delta(s_t \TRANS{\prod_{j \in \AG} a^t_{j}} s_{t+1} ) > 0$.
\end{definition}

\begin{definition}
\label{def:hist}
Given a \pomas\ $\MMM=(\AG, \GGG, s_0, \ASP, L, \{\obs_i\}_{i \in \AG})$,
a \emph{history} is a finite path starting with $s_0$,
the set of histories in $\MMM$ is written as $\HIST(\MMM)$
and the set of histories in $\MMM$ starting with history $h$ is written as $\HIST(\MMM,h)$.
For any agent $i \in \AG$, and two histories $h$ and $h'$,
we say $h$ and $h'$ are observationally equivalent to each other from $i$'s view, 
denoted by $h \sim_i h'$, iff
$\obs_i(h) = \obs_i(h')$.
\end{definition}

\begin{example}
\label{eg:str}
Consider the second scenario proposed in Example~\ref{eg:model}. The observer's information of knowledge obtained from her observation might influence her decision on voting. 
Assume the observer (voter 3) is not able to see whom other voters have voted,
but she can see how many ballots each candidate has received as specified in Table~\ref{tbl:eg-actions} and thus she can indicate the dominant candidate so far.
A basic strategy to reflect such an influence is that 
she will vote the dominant candidate if there is one, 
otherwise 
she will vote the candidates under her preferred distribution.
\end{example}


\section{Observability Specification}
\label{sec:logic}
This section studies the problem of formally specifying observability of an agent on system behaviours modelled in \pomas. 

\subsection{Observability and opacity}
Given a property $\varphi$ and an observation function $\obs_i$ of an agent $i \in \AG$, 
we are interested in quantitatively expressing the observability of the agent 
that a set of agents has a strategy to  enforce the property $\varphi$.
The property can be viewed as a predicate, \ie a set of execution paths for which it holds.
The concept of \emph{Opacity}~\cite{Mazare04a} provides an intuitive approach for this task via distinguishing the observed behaviour and the original one. 
Intuitively, a property $\varphi$ is opaque (not observable), 
provided that for every behaviour (say path $\rho$) satisfying $\varphi$ 
there is another behaviour (say path $\rho'$), not satisfying $\varphi$, 
such that $\rho$ and $\rho'$ are observationally equivalent. 
So the observer is not able to determine whether the property 
in a given path of the system is satisfied or not.
More precisely, opacity specifies whether an agent can establish 
a property $\varphi$, enforced by a strategy of a coalition $A$ of agents, 
at some specific state(s) of the executions of the system, 
according to her observation on the system behaviours.
We use $[\![\varphi]\!]$ to denote the set of paths satisfying property $\varphi$.
\begin{definition}
\label{def:opacity}
Let $\MMM=(\AG, \GGG, s_0, \ASP, L, \{\obs_i\}_{i \in \AG})$. Given a predicate $\varphi$ over $\RUN{s_0}{\GGG}$,
we say $\varphi$ is \emph{opaque} \wrt $\obs_i$ if
for every path $\rho \in [\![\varphi]\!]$, 
there is a path $\rho' \in \bar{[\![\varphi]\!]}$ s.t.~$\obs(\rho)=\obs(\rho')$, 
\ie all paths satisfying $\varphi$ are covered by paths in 
$\bar{[\![\varphi]\!]}$: 
$\obs_i([\![\varphi]\!]) \subseteq \obs_i(\bar{[\![\varphi]\!]})$ under $\obs_i$,
where $\bar{[\![\varphi]\!]} \triangleq \RUN{s_0}{\GGG} \setminus [\![\varphi]\!]$.
\end{definition} 

\subsection{The logic \opacpctl}
To express the observability of an agent, we would consider the transparent paths, \ie behaviours observable (non-opaque) to her. 
The level of observability can be considered as the degree of transparency of 
the property enforced by the strategy of a coalition, 
which can be measured by calculating the probability of the transparent paths 
satisfying the property.
We now present \opacpctl, an extension of 
probabilistic alternating-time temporal logic (\patl)~\cite{ChenL07},
that characterises agents' quantified ability to enforce temporal properties.
The key additions of \opacpctl\ include an \emph{observability operator} 
and a \emph{probabilistic (observability) operator}.

\begin{definition}
\label{def:logic}
Let $\MMM = (\AG, \GGG, s_0, \ASP, L, \{\obs_i\}_{i \in \AAA})$.
The syntax of \opacpctl\ includes three classes of formulae: 
\emph{state and path formulae}, and
\emph{observability formulae} 
ranged over by $\phi$, $\psi$ and $\Phi$, respectively.
\begin{GRAMMAR}
  \phi
     &::=&
  a 
     \VERTICAL
  \neg \phi
     \VERTICAL
  \phi \land \phi
     \VERTICAL
  \P_{\bowtie p} \, \langle \AAA \rangle \lbrack {\psi} \rbrack
     \VERTICAL
  \D_{\bowtie p} \, \langle \AAA \rangle \lbrack {\Phi} \rbrack
     \\
  \psi
     &::=&
  \X \phi 
     \VERTICAL
  \phi  \U \phi 
     \VERTICAL
     \phi  \R \phi
    \VERTICAL
  \neg \psi
     \VERTICAL
  \psi \land \psi      
     \\
  \Phi
     &::=&
  \opac_i \, \lbrack \psi \rbrack
    \VERTICAL
  \neg \Phi
     \VERTICAL
  \Phi \land \Phi
\end{GRAMMAR}
\noindent
where $a \in \ASP$ is an \emph{atomic proposition},
$\AAA \subseteq \AG$ is a set of agents,
$\langle \AAA \rangle$ is the strategy quantifier,
$\langle \AAA \rangle \lbrack \psi \rbrack$ expresses the property that 
coalition $\AAA$ has a strategy to enforce $\psi$,
$i \in \AAA \subseteq \AG$ is an agent,
${\bowtie} \in \{\le, <, \ge, >\}$, 
$p \in \lbrack 0,1 \rbrack$ is a probability bound.  
\end{definition}
Note that \opacpctl\ formula is defined relative to a state,
path formulae are only allowed inside the observability operator $\opac_i\, \lbrack \cdot \rbrack$ 
and the probabilistic operator  $\P_{\bowtie p}\, \langle \AAA \rangle \lbrack \cdot \rbrack$.
The formula $\P_{\bowtie p} \ \langle \AAA \rangle \lbrack \psi \rbrack$ expresses that $\AAA$ has a strategy
such that the probability of satisfying path formula $\psi$ is $\bowtie p$, when the strategy is followed.
The observability formula $\opac_i \, \lbrack \psi \rbrack$
expresses the property of behaviours satisfying $\lbrack \psi \rbrack$ are observable to agent $i$.
Intuitively, it is satisfied if for each path $\rho$ satisfying $\psi$ one cannot find a path $\rho'$ violating $\psi$ such that $\rho$ and $\rho'$ observationally equivalent to each other - from agent $i$'s view.
This operator would allow us to reason about the observability of agent $i$ on system behaviours to enforce the property $\psi$.
The quantitative observability formula $\D_{\bowtie p} ( \langle \AAA \rangle \lbrack \Phi \rbrack)$ expresses that $\AAA$ has a strategy $\pi_{\AAA}$ such that the degree of the observability enforcing path property considered in $\Phi$ is $\bowtie p$.
%
$ \RUN{s,\pi_{\AAA}}{\GGG}$ is used to denote the set of all paths of $\GGG$ starting from $s$ and consistent with $\pi_{\AAA}$.

\begin{definition}
Let $\MMM = (\AG, \GGG, s_0, \ASP, L, \{\obs_i\}_{i \in \AG})$.
Semantics for \opacpctl\ include three satisfaction relations 
regarding the three notions of formulae (state, path, observability formulae). 

For a state $s \in S$ of $\GGG$, the \emph{satisfaction relation}
$s \models_{\MMM} \phi$ for state formulae denotes ``$s$ satisfies $\phi$'':

\begin{itemize}


  \item $s \models_{\MMM} a$ iff $a \in L(s)$.

  \item $s \models_{\MMM} \neg\phi$ iff $s \not \models_{\MMM} \phi$.

  \item $s \models_{\MMM} \phi \land \phi'$ iff 
  	$s \models_{\MMM} \phi$ and  $s \models_{\MMM}  \phi'$.

 \item $s \models_{\MMM} \P_{\bowtie p} \langle \AAA \rangle \lbrack \psi \rbrack$ iff 
	$\exists \pi_{\AAA}$,  the probability of the consistent paths with 
	$\pi_{\AAA}$ over the set $\AAA$, 
	from state $s$, that $\psi$ is true, satisfies $\bowtie p$, \ie
	$\prob(s, [\![\langle \AAA \rangle \lbrack \psi \rbrack ]\!]) \bowtie p$,
	where: 
	$\prob(s, [\![ \langle \AAA \rangle \lbrack \psi \rbrack ]\!])  
	= \PPP_s ([\![\langle \AAA \rangle \lbrack \psi \rbrack ]\!])    
	= \PPP_s\{\rho \in \RUN{s,\pi_{\AAA}}{\GGG} \mid 
	\rho \models \psi \}$.
  
 \item $s \models_{\MMM} \D_{\bowtie p} \langle \AAA \rangle \lbrack \Phi \rbrack$ iff
 from state $s$, 
 the probability of outgoing observable paths enforced by $\Phi$
 that are consistent with $\pi_{\AAA}$ of a coalition $\AAA$, 
 satisfies $\bowtie p$:
 $\prob(s, [\![\langle \AAA \rangle \Phi]\!]) \bowtie p$,
 where for the case of 
 $\Phi= \opac_i \lbrack \psi \rbrack$ and 
 $\Phi'= \opac_j \lbrack \psi' \rbrack$: 
\begin{footnotesize}
\begin{eqnarray*}
\prob(s, [\![\langle \AAA \rangle \Phi]\!]) 
 =& \hspace{-2mm} \PPP_s ( [\![\langle \AAA \rangle \lbrack \psi \rbrack]\!] \setminus \obs^{-1}_i(\obs_i([\![\neg \psi ]\!])) ) 
 \\
 \prob(s, [\![\neg \Phi]\!]) 
 =& \hspace{-2mm} \PPP_s( [\![\langle \AAA \rangle \lbrack \psi \rbrack]\!] \cap \obs^{-1}_i(\obs_i([\![\neg \psi ]\!])) )
 \\
 \prob(s, [\![\Phi \land \Phi' ]\!]) 
 =& \hspace{-2mm} \PPP_s ( ([\![\langle \AAA \rangle \lbrack \psi \rbrack ]\!] \setminus \obs^{-1}_i(\obs_i([\![\neg \psi ]\!] ))) 
\\
  & \hspace{-3mm} \cap  ([\![\langle \AAA \rangle \lbrack \psi' \rbrack ]\!]
 			\setminus \obs^{-1}_j(\obs_j([\![ \neg \psi']\!]))) ).
\end{eqnarray*}
\end{footnotesize}
\end{itemize}

For a path $\rho$ of $\GGG$, we define:

\begin{itemize}

\item $\rho \models_{\MMM} \X \phi$ iff $\rho_s(1) \models \phi$.
  
\item $\rho \models_{\MMM} \phi \U \phi'$ iff there exists $i \in
  \NAT$ s.t.~$\rho_s(i) \models_{\MMM} \phi'$ and $\rho_s(j) \models_{\MMM} \phi$  for all $j < i$.
    
\item $\rho \models_{\MMM} \phi \R \phi'$ iff for all $i \in \NAT$ at least one of the following is true: i) $\rho_s(i) \models_{\MMM} \phi'$, ii) $\rho_s(j) \models_{\MMM} \phi$  for some $j < i$.

  
  \item $\rho \models_{\MMM} \neg\psi$ iff $\rho \not \models_{\MMM} \psi$.

  \item $\rho \models_{\MMM} \psi \land \psi'$ iff $\rho \models_{\MMM} \psi$ and  $\rho \models_{\MMM}  \psi'$.
  
\end{itemize}

\noindent
Finally, for $s \models_{\MMM} \Phi$, we define observability formulae $\Phi$:
\begin{itemize}

  \item $s \models_{\MMM} \opac_i \, \lbrack \psi \rbrack$ iff 
   for each path $\rho \in \RUN{s}{\GGG}$ 
   s.t.~$\rho\models_{\MMM} \psi$,
    and for all $\rho' \in \RUN{s}{\GGG}$ 
    s.t.~$\rho' \not\models_{\MMM} \psi$: 
    $\obs_i(\rho) \ne \obs_i(\rho')$.

 \item $s \models \neg\Phi$ iff $s \not \models \Phi$, \ie
    for each path $\rho \in \RUN{s}{\GGG}$ s.t.~$\rho\models_{\MMM} \psi$,
     there exists a path $\rho' \in \RUN{s}{\GGG}$ s.t.~$\rho' \not\models_{\MMM} \psi$: 
     $\obs_i(\rho) = \obs_i(\rho')$.

 \item $s \models \Phi \land \Phi'$ iff 
 $s \models_{\MMM} \Phi$ and $s \models_{\MMM} \Phi'$.
     
\end{itemize}
\end{definition}

\begin{example}
\label{eg:logic}
Consider the model in Example~\ref{eg:model}. 
Assume we are interested in analysing the observability of voter $2$ regarding her observation function in Table~\ref{tbl:eg-actions}.
Consider the property \emph{``eventually candidate $X$ wins"}, 
\ie $\psi = \F (cx>cy \land o=2)$,
where $cx$ and $cy$ denotes the final ballots $X$ and $Y$ received,
and $o=2$ indicates the state of voting process being closed.
The operator $\F \phi$ is defined as ``$\true ~ \U ~ \phi$'', 
so the probabilistic observability property is specified as:
$
\D_{\le p} \langle 1,2,3 \rangle \lbrack \opac_2   \lbrack \psi \rbrack\rbrack
$,
where $p=\frac{1}{3}$ is a probability threshold.
It is easy to notice that,
the above formula would return $\true$, since (``wait'' actions have been omitted here for simplifying the expression without introducing any confusion):
{\footnotesize
\begin{eqnarray*}
\TR{[\![\psi ]\!]} =& \hspace{-3mm}
\{\frac{1}{12}. \open_0V^X_1V^X_2V^X_3\close_0, 
\frac{1}{6}. \open_0V^X_1V^X_2V^Y_3\close_0, \\
& \frac{1}{12}. \open_0V^X_1V^Y_2V^X_3\close_0,
\frac{1}{6}. \open_0V^Y_1V^X_2V^X_3\close_0\}
\end{eqnarray*}}
\noindent
and thus,
{\footnotesize \[\TR{[\![ \opac_2 \lbrack \psi \rbrack]\!]} \!=\!
\{\frac{1}{12}. \open_0V^X_1V^X_2V^X_3\close_0, 
\frac{1}{6}. \open_0V^X_1V^X_2V^Y_3\close_0 \},\]}
\noindent
this is because under the observation function specified in Table~\ref{tbl:eg-actions}, 
traces {\footnotesize $\open_0V^X_1V^Y_2V^X_3\close_0$} 
and {\footnotesize $\open_0V^Y_1V^X_2V^X_3 \close_0$} 
are covered by violating $\psi$ traces 
{\footnotesize $\open_0V^X_1V^Y_2V^Y_3\close_0$} 
and {\footnotesize $\open_0V^Y_1V^X_2V^Y_3 \close_0$} respectively from voter 2's view:
{\footnotesize
\begin{eqnarray*}
\obs_2(\open_0V^X_1V^Y_2V^X_3\close_0) \!=\! \obs_2(\open_0V^X_1V^Y_2V^Y_3\close_0) \!=\! \open_0Y_2\close_0,\\
\obs_2(\open_0V^Y_1V^X_2V^X_3\close_0) \!=\! \obs_2(\open_0V^Y_1V^X_2V^Y_3\close_0) \!=\! \open_0X_2\close_0.
\end{eqnarray*}
}
%
Therefore, 
{\footnotesize
\begin{eqnarray*}
 && \hspace{-3mm} 
\prob([\![ \langle 1,2,3 \rangle\opac_2 \lbrack \psi \rbrack]\!])  \\
&=& \hspace{-3mm} \prob(\{\frac{1}{12}. \open_0V^X_1V^X_2V^X_3\close_0, 
\frac{1}{6}. \open_0V^X_1V^X_2V^Y_3\close_0\})\! =\! \frac{1}{4} 
\end{eqnarray*}}

\noindent
which is less than $p=\frac{1}{3}$ 
and thus $\D_{\le p} \langle 1,2,3 \rangle  \lbrack \opac_2 \lbrack \psi \rbrack\rbrack $ returns $\true$.
\end{example}

\section{Verification of Observability Properties}
\label{sec:veri}

Intuitively, verification of probabilistic observability answers the question ``to which degree the system is observable to an agent $i  \in \AG$?'', relative to a task expressed as property $\lbrack \psi \rbrack$ following the strategy of a coalition $\AAA \subseteq \AG$, and the observation function of the agent $\obs_i$.
Since \opacpctl\ is a branching time logic, 
the overall approach is to
recursively compute the satisfaction set $\SAT(\phi)$ of states satisfying formula $\phi$ over the structure of the formula.
%

For the \textit{propositional logic fragment} of \opacpctl, 
the computation of this set for atomic propositions and
logical connectives follows the conventional CTL model checking~\cite{Baier2008} and is sketched below:

 \noindent{(1)}  
Convert the $\opacpctl$ formulae in a positive normal form, that is, formulae built by the basic modalities $\OPAC{\X \phi}$, $\OPAC{\phi  \U \phi'}$, and $\OPAC{\phi \R \phi'}$, and successively pushing negations inside the formula at hand: $\neg \true \leadsto \false$, $\neg \false \leadsto \true$, $\neg\neg \phi \leadsto \phi$, $\neg(\phi \land \phi') \leadsto \neg \phi \lor \neg \phi'$, $\neg(\phi \lor \phi') \leadsto \neg \phi \land \neg \phi'$, $\neg \X \phi \leadsto \X \neg \phi$, $\neg(\phi \U \phi') \leadsto \neg\phi \R \neg\phi'$, $\neg(\phi \R \phi') \leadsto \neg\phi \U \neg\phi'$;

 \noindent{(2)}
Recursively compute the satisfaction sets $\SAT(\phi') = \{s \in S \mid s \models \phi'\}$ for all state subformulae $\phi'$ of $\phi$: the computation carries out a bottom-up traversal of the parse tree of the state formula $\phi$ starting from the leafs of the parse tree and completing at the root of the tree which corresponds to $\phi$, where the nodes of the parse tree represent the subformulae of $\phi$ and the leafs represent an atomic proposition $\alpha \in \ASP$ or $\true$ or $\false$. 
All inner nodes are labelled with an operator. For positive normal form formulae, the labels of the inner nodes are $\neg$, $\land$, $\OPAC{\X}$, $\OPAC{\U}$, $\OPAC{\R}$.
At each inner node, the results of the computations of its children are used and combined to build the states of its associated subformula. 
In particular, satisfaction sets for the propositional logic fragment state formula are given as follows: 
\vspace{-1.5mm}
   \begin{itemize}
	 \item $\SAT(\true) = S$,
	 
     \item $\SAT(\alpha) = \{t \in S \mid \alpha \in \eta(t)\}$,

     \item $\SAT(\neg \phi) = S \setminus \SAT(\phi)$,

     \item $\SAT(\phi \land \phi') = \SAT(\phi) \cap \SAT(\phi')$,
     
     

     \end{itemize}
 \noindent{(3)}
Check whether $s \in \SAT(\phi)$.

For the treatment of subformulae of the form $\phi= \P_{\bowtie p} \langle \AAA \rangle \lbrack {\psi} \rbrack$,
in order to determine whether $s \in \SAT(\phi)$, 
the probability of consistent paths with $\pi_{\AAA}$ under coalition $\AAA$ for behaviour specified by $\psi$, 
\ie $\prob(s \models_{\MMM} \langle \AAA \rangle \lbrack {\psi} \rbrack)$, needs to be established, then:
\[\SAT(\P_{\bowtie p} \langle \AAA \rangle \lbrack {\psi} \rbrack) = \{s \in S \mid \prob( s \models_{\MMM}  \langle \AAA \rangle \lbrack {\psi} \rbrack) \bowtie p \}\]
The computation of the probability can follow the conventional PATL model checking algorithms, \eg~\cite{ChenL07}.
%

We now focus on the treatment state formulae of the form 
$\D_{\bowtie p} \lbrack \opac_i\langle \AAA \rangle \psi \rbrack$.
The problem reduces to computing the probability of observable paths that are satisfying property $\psi$ and consistent with strategies of coalition $\AAA$, from agent $i$'s view.
%
\begin{definition}
\label{def:mc}
Given a \pomas\ $\MMM=(\GGG, s_0, \AG, \ASP, \{\obs_i\}_{i \in \AG})$, 
a task in property $\psi$ required to be completed under 
a strategy $\pi_{\AAA}$ of a coalition $\AAA \in \AG$, 
the probabilistic verification problem of observability property 
is to decide whether
$s_0 \models_{\MMM} \D_{\bowtie p} ( \opac_i \ \langle \AAA \rangle \lbrack \psi \rbrack )$, \ie
\[
\PPP_{s_0} ( [\![\langle \AAA \rangle \lbrack \psi \rbrack ]\!]\setminus \obs^{-1}_i(\obs_i([\![\langle \AAA \rangle \lbrack \neg \psi \rbrack]\!])) ) ~\bowtie~ p.
\]
\end{definition}
Therefore, we focus on computing $\PPP_{s_0} ( \langle \AAA \rangle \lbrack \psi \rbrack \setminus \obs^{-1}_i(\obs_i(\langle \AAA \rangle \lbrack \neg\psi \rbrack)) )$ for a given \pomas\ $\MMM$ and coalition $\AAA$.
We assume that the available actions of agent $i \in \AG$ of $\MMM$ in state $s$ are $\{a_{i,1}, \dots, a_{i, k_i}\}$.
The brief procedure for checking 
$s \models_{\MMM} \D_{\bowtie p} ( \opac_i \ \langle \AAA \rangle \lbrack \psi \rbrack)$ is sketched as follows.
\begin{itemize}
\item Find all consistent paths $\Pi$ and the corresponding traces $\Lambda$, represented in regular-expression-like format (denoted by $\reg(\cdot)$), satisfying $\psi$ under mixed strategy $\pi_{\AAA}$ of coalition $\AAA$, denoted by: 
\[
\Pi=\{\reg(\rho_{\pi_{\AAA}}) \mid \rho_{\pi_{\AAA}} \models_{\MMM} \psi\}
\qquad 
\Lambda = \{\ERASE{\rho} \mid  \rho \in \Pi\}.
\]
\item Find all consistent paths $\Pi'$ and the corresponding traces $\Lambda$, represented in regular-expression-like format (denoted by $\reg(\cdot)$), violating $\psi$ under mixed strategy $\pi_{\AAA}$ of coalition $\AAA$: 
\[
\Pi'=\{\reg(\rho'_{\pi_{\AAA}}) \mid \rho_{\pi_{\AAA}} \not\models_{\MMM} \psi\}
\qquad 
\Lambda' = \{\ERASE{\rho'}  \mid  \rho' \in \Pi'\}.
\]
\item Find all $\psi$-opaque traces: 
\[\Lambda'' = \{\lambda'' \mid \lambda'' \in \Lambda \land \exists \lambda' \in \Lambda'.(\obs_i(\lambda')=\obs_i(\lambda''))\}.\]
\item Compute the probability of $\psi$-observable traces:
\[
d = \PPP_{s_0} ([\![ \opac_i \langle \AAA \rangle \lbrack \psi \rbrack ]\!]) = \sum_{\xi \in (\Lambda \setminus \Lambda'')} \prob(\xi).
\]
\item Return true if $d \bowtie p$, return false otherwise.
\end{itemize}
We present the detailed procedure of computing the probability of $\psi$-observable traces starting at $s$ under mixed strategy $\pi_{\AAA}$ of coalition $\AAA$ from the observation of $i \in \AG$, in Algorithm~\ref{algo:veri}. 
Algorithm~\ref{algo:compU} computes a set of regular-expression-like formatted paths satisfying $\phi \U \phi'$. 
Similarly, an algorithm can be proposed to compute a set of regular-expression-like formatted paths satisfying $\phi \R \phi'$.
We can thus compute all regular-expression-like formatted paths $\Pi$ (and $\Pi'$) starting from $s$ and satisfying (and violating) $\psi$ and consistent with mixed strategy $\pi_{\AAA}$.

\begin{algorithm}[!t]
\caption{Computing the probability of $\psi$-observable consistent traces under $\pi_{\AAA}$ from $i$'s view - $\D(\langle \AAA \rangle \opac_i \lbrack \psi \rbrack)$.} 
 \label{algo:veri}
\begin{footnotesize}
 \SetAlgoLined
  \KwData{$\MMM, s, i, A, \psi$}
  \KwResult{The probability $\D(\langle \AAA \rangle \opac_i \lbrack \psi \rbrack)$} 
 \Switch{$\psi$} {
	\textbf{case} $\X \phi$: \ \  $\SAT(\psi) \leftarrow \cup \{s \TRANS{\alpha} s' \mid \POST(s)=s'  \land s' \in \SAT(\phi)\}$, \\ 
	$~~\SAT(\neg\psi) \leftarrow \cup \{s \TRANS{\alpha} s' \mid \POST(s)=s'  \land s' \in \SAT(\neg\phi)\}$\;
	\textbf{case} $\phi \U \phi'$: $\SAT(\psi) \leftarrow \textbf{compU}(\MMM, s, \phi, \phi')$, \\ 
	$\qquad \qquad \quad  \SAT(\neg\psi) \leftarrow \textbf{compR}(\MMM, s, \neg\phi, \neg\phi')$\;
	\textbf{case} $\phi \R \phi'$: $\SAT(\psi) \leftarrow \textbf{compR}(\MMM, s, \phi, \phi')$, \\ 
	$\qquad \qquad \quad \SAT(\neg\psi) \leftarrow \textbf{compU}(\MMM, s, \neg\phi, \neg\phi')$\;
	}  
	$p\Lambda \leftarrow  \{p\lambda \mid p\lambda.tr \leftarrow \lambda \land p\lambda.pr \leftarrow \prob(\lambda)  \land \lambda=tr(\rho) \ \text{for} \ \rho \in \RUN{s,\pi_{\AAA}}{\GGG} \land \rho \in \SAT(\psi)\}$\;
	$p\Lambda' \leftarrow \{p\lambda \mid p\lambda.tr \leftarrow \lambda \land p\lambda.pr \leftarrow \prob(\lambda) \land \lambda=tr(\rho) \ \text{for} \ \rho  \in \RUN{s,\pi_{\AAA}}{\GGG} \land \rho \in \SAT(\neg\psi)\}$\;
   $p\Lambda'' = \{\}$\;
  \For{each $p\lambda \in p\Lambda$}
  {
		\For {each $p\lambda' \in p\Lambda'$}
		{
	  	\If{$\obs_i(p\lambda.tr) \subseteq \obs_i(p\lambda'.tr)$}
	  	{ 
	    	$p\Lambda'' \leftarrow p\Lambda'' \cup \{p\lambda\}$; break\;
	  	}
	  }
  }
  ${p\Lambda}_{\opac} \leftarrow p\Lambda \setminus p\Lambda''$; \tcc*[f]{\scriptsize observable traces} \\
  $d=\sum_{\lambda \in p\Lambda_{\opac}} \lambda.pr$\;
  \Return{$d$}.

\end{footnotesize}
\end{algorithm}

%


\begin{algorithm}[!ht]
\caption{Computing $\SAT(\phi \U \phi')$: \bf{compU}($\MMM, s, \phi ,\phi'$)}
 \label{algo:compU}
\begin{footnotesize}
 \SetAlgoLined
  \KwData{$\MMM, s, \phi, \phi'$}
  \KwResult{Regular-expression-like formatted paths satisfying $\phi \U \phi'$}
    $\Pi \leftarrow \{\}$; 
    $i \leftarrow 0$ \;
	\For{each $t_i \in \SAT(\phi')$}
	{
	    $T_i \leftarrow \{t_i\}; \ \Pi_i \leftarrow \{\pi \mid \pi(0)=t_i\}$\;
	    \While{$\{s_j \in \SAT(\phi) \setminus (T_i \cup \SAT(\phi')) \mid \POST(s_j) \cap T_i \ne \emptyset\} \ne \emptyset$}
	    {
	    	let $s_j \in \{s_j \in \SAT(\phi) \setminus (T_i \cup \SAT(\phi')) \mid \POST(s_j) \cap T_i \ne \emptyset\}$\;
	    	\If{$s_j \in \POST(s_j) \cap T_i$}
	    	{
	    		/* There is a self-loop: wrap it with a star and concatenate paths starting from a state in $\POST(s_j) \cap T_i$ */ \\
	    		\For {each $\pi' \in \Pi_i$ s.t. $\pi'(0) \in \POST(s_j) \cap T_i$}
	    		{
	    			$\Pi_i \leftarrow \Pi_i \cup \{(s_j \TRANS{\alpha} s_j)^* + \pi'\lbrack 1... \rbrack \}$\;
	    		}
	    	}	    
	    	\For {each: $q_1\in \POST(s_j) \cap T_i, q_2 \in \POST(q_1) \cap T_i, \dots, q_n \in \POST(q_{n-1}) \cap T_i$ s.t. $\POST(q_{n}) \cap T_i = \emptyset$}
	    	{
	    		\If{$\PRE(s_j) \not\in \{q_1, q_2, \dots q_n\}$}
	    		{
	    			\For {each $\pi' \in \Pi_i$ s.t. $\pi'(0) \in \POST(s_j) \cap T_i$}
	    			{
	    			$\Pi_i \leftarrow \Pi_i \cup \{s_j \TRANS{\alpha} q_1 + \pi'\lbrack 1... \rbrack \}$\;
	    			}
	    		}
	    		\ElseIf {$\PRE(s_j) = q_n \land s_j \in \POST(q_{n})$}
	    		{
	    			/* There is a cycle, wrap it with a star and concatenate paths starting from a state in $\POST(s_j) \cap T_i$ */ \\
	    		
	    			\For {each $\pi' \in \Pi_i$ s.t. $\pi'(0) \in \POST(s_j) \cap T_i$}
	    			{
	    			$\Pi_i \leftarrow \Pi_i \cup \{(s_j \TRANS{\alpha_1} q_1 \TRANS{\alpha_2} \dots \TRANS{\alpha_{n}} \PRE(s_j) \TRANS{\alpha_{n+1}} s_j)^* + \pi'\lbrack 1... \rbrack\}$\;
	    			}
	    		}
	    	}
	    	$T_i \leftarrow T_i \cup \{s_j\}$\;
	    	}
	    $\Pi \leftarrow\Pi \cup \Pi_i$ \;
	    $i \leftarrow i+1$\;
	 }
  \Return{$\Pi$}.
 \end{footnotesize}
\end{algorithm}

\smallskip\noindent
{\bf Soundness.} 
Given a \pomas\ $\MMM$, a probability threshold $p$, and a task specified in $\psi$ to be completed:
\[
s_0 \models_{\MMM} \D_{\bowtie p} \langle \AAA \rangle  \lbrack \opac_i \lbrack \psi \rbrack\rbrack
\qquad \text{iff} \qquad
\PPP ([\![\langle \AAA \rangle \lbrack \opac_i  \lbrack \psi \rbrack\rbrack ]\!]) \bowtie p.\]
The satisfaction relation of 
$\D_{\bowtie p} \langle \AAA \rangle \lbrack \opac_i\, \lbrack \psi \rbrack\rbrack$ 
and the computation of the probability of $\psi$-observable consistent traces under mixed strategy $\pi_{\AAA}$ of coalition $\AAA$ from observer $i$'s view is described in Algorithm~\ref{algo:veri}. 
The algorithm will terminate since $\SAT(\psi)$ are processed and computed
as a set of regular-expression-like formatted traces satisfying $\psi$ (such as Algorithm~2 presented in the technical appendix).
Probability of such a trace is calculated by multiplication of the probability of each transition label for non-cycle part, and multiplication of $p/(1-p)$ for a cycle with probability $p$.

\smallskip\noindent
{\bf Complexity.}
The worst case of checking satisfaction of the observability formula, specified in Algorithm~\ref{algo:veri}, is EXPSPACE in general. The formula of observability is essentially in the form of $\forall \forall$, the algorithm traverses all consistent traces under mixed strategy $\pi_{\AAA}$ satisfying $\psi$ and all traces of those violating $\psi$, and conducts observation equivalence comparison. So the worst case complexity here follows the complexity of the hyper property model checking problem with two quantifier ($\forall$) alternations, and thus EXPSPACE. We hypothesise the time complexity of checking satisfaction of \opacpctl formula is exponential to the size of the POMAS, and doubly exponential in the size of the formula itself, similar to model checking HyperLTL~\cite{Clarkson14}. 
If all pairs of traces are evaluated in parallel, the evaluation of each individual pair corresponds to the evaluation of an LTL formula over a single trace, which can be done in polylogarithmic time on a parallel computer with a polynomial number of processors~\cite{Bonakdarpour21}.

\begin{example}
The proposed work has been implemented on top of PRISM,
which allows to specify properties which evaluate to a value using, e.g., 
$\D_{= ?} \langle A \rangle \lbrack \opac_i \lbrack \psi \rbrack \rbrack$.
The result of Example~\ref{eg:logic} can be automatically produced below, which meets our calculation by hand.
{\scriptsize
\vspace{-2mm}
\begin{verbatim}
Result: 0.25.{
    0.08333333333333333:vX1vX2vX3cl0->X2cl0,
    0.16666666666666666:vX1vX2vY3cl0->X2cl0
} (value in the initial state)
\end{verbatim}}
\end{example}



\section{Implementation and Examples}
\label{sec:impl}
A prototype tool for specifying and verifying the observability problem in MASs has been built on the top of the PRISM model checker~\cite{KNP11}. 
Models are described in an extension of the PRISM modelling language with observations and transition labels,
the new model type is denoted as ``\texttt{pomas}''. 
Properties are described in an extension of the PRISM's property specification language with the observability operator. The tool and examples are available from~\cite{impl}.


\smallskip\noindent
{\bf Example: a simple supply chain.}
Nowadays supply chain is a core part of businesses concerned with transporting products between different parties such as customers, retailers, coordinators, delivery services, and suppliers. Agents of those parties communicate with each other for buying and selling items. Suppliers compete with each others to obtain more jobs and profit, they might partially observe the procedure of the supply chain and try to induce commercial information. Customer might partially observe the pipeline of the supply chain, and try to learn information about the origin of the products. 
Such a scenario can be naturally modelled as a POMAS,
and we are interested in studying the quantified observability by agents, 
which may cause information flow and affect future decision-making. 

\begin{figure}[!t]
    \centering
\scalebox{0.75}
{\begin{tikzpicture}[->,>=stealth',shorten >=1pt,auto,node distance=2.5cm, scale = 1, transform shape]

  \node[state] (A) {sup1};
  \node[state] (C) [below right of=A] {coord};
  \node[state] (B) [below left of=C] {sup2};
  \node[state] (D) [right of=C] {retailer};
  \node[state] (E) [right of=D] {customer};

   \path[every node/.style={font=\sffamily\small,
   		fill=white,inner sep=1pt}]
     (A) edge [bend right=20] node[left=1mm] {res1} (C)
     (B) edge [bend left=20] node[left=1mm] {res2} (C)
     (C) edge [bend right=20] node [right=1mm]{1/2.req1} (A)
     (C) edge [bend left=20] node [right=1mm]{1/2.req2} (B)
     (C) edge [bend left=20] node [above=1mm]{decision} (D)
     (D) edge [bend left=20] node [below=1mm]{ordr} (C)
     (D) edge [bend left=20] node [above=1mm]{delivery} (E)
     (E) edge [bend left=20] node [below=1mm]{ordc} (D)
;
\end{tikzpicture}}
\vspace{-3mm}
    \caption{A simple supply chain.}
    \label{fig:eg-sup}
\vspace{-3mm}
\end{figure}
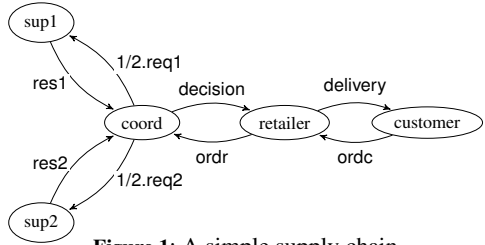 

To illustrate our framework and its implementation, 
we consider a basic commercial supply chain shown in Fig.~\ref{fig:eg-sup} as an example.
Assume there are a number of agents in the system:
1) customer: buying products (denoted by \texttt{ordc}) from the retailer;
2) retailer: requesting to order products (denoted by \texttt{ordr}) from suppliers through the coordinator; 
3) coord: the coordinator, processing requests/orders from the retailer, sending requests to
and receiving response from suppliers, making decisions such as which supplier provides products, returning decision to the retailers;
4) sup$i$: the $i^{\it th}$ supplier, receiving requests from/responding availability to the coordinator. 
The agents and their observability are given as follows:
action \texttt{ordc} is hidden to \textit{supi} and is observed as \texttt{Ordc} from the rest of the agents' view;
action \texttt{ordr} is hidden to \textit{customer} and is observed as \texttt{Ordr} from the rest of the agents' view;
action \texttt{reqi}, denoting \textit{coord} sending requests to \textit{supi}, is hidden to \textit{customer \& retailer}, is observed as \texttt{Req} to \textit{supi}, and is observed as \texttt{Reqi} to \textit{coord};
action \texttt{resi}, denoting \textit{supi} responding to \textit{coord}, is hidden to \textit{customer \& retailer}, is observed as \texttt{Res} to \textit{supi}, and is observed as \texttt{Resi} to \textit{coord};
action \texttt{decisioni}, denoting \textit{coord} deciding \textit{supi} to provide the products, is hidden to \textit{customer \& retailer}, is observed as \texttt{Dec} to \textit{supi}, and is observed as \texttt{Deci} to \textit{coord};
action \texttt{delivery} is observed \texttt{Dlv} to all of the agents.

Let $A$ denote the set of agents defined above. We could ask questions such as
``what is the degree of the observability by \textit{sup1} if the product is successfully delivered to the customer
but the supplier is not \textit{sup1}?'', specified as $\P=_? \langle A \rangle \ \lbrack\opac_{\textit{sup1}} \  
\lbrack \F \ (\texttt{dec} != 1 \ \& \ \texttt{dlv}=1) \rbrack\rbrack$,
%
%
where \texttt{dec} is the variable defined in the module to specify the decision made by the coordinator: \texttt{dec=i} denotes supplier $i$ will provide the requested products, 
\texttt{dlv} is the variable defined in the module to specify the status of product delivery: \texttt{dlv=1} denotes the product has been successfully delivered to the customer. 
The result generated by the tool is presented as follows:
\begin{scriptsize}
\vspace{-2mm}
\begin{Verbatim}
Result: 0.5.{
  0.25:ordcordrreq1res1decision2delivery->OrdrReqResDecDlv,
  0.25:ordcordrreq2res2decision2delivery->OrdrReqResDecDlv
} (value in the initial state)                
\end{Verbatim}
\end{scriptsize}
This meets our intuition, since the listed two traces satisfying $\F (\texttt{dec}!=1\& \texttt{dlv}=1) $
are not covered by traces violating the property,
are thus observable to \textit{sup1}.
If we ask question ``what is the degree of the observability by \textit{customer} if the product is successfully delivered to him but the supplier is not \textit{sup1}?'', which can be specified as $\P=_?  \langle A \rangle \ \lbrack\opac_{\textit{customer} } \ 
\lbrack \F \ (\texttt{dec} != 1 \ \& \ \texttt{dlv}=1) \rbrack\rbrack$. 
%
The result generated by the tool would be:
\begin{scriptsize}
\vspace{-2mm}
\begin{Verbatim}
Result: 0.0.{
} (value in the initial state)         
\end{Verbatim}
\end{scriptsize}

\begin{figure}[!t]
    \centering
 \scalebox{0.75}
{\begin{tikzpicture}[->,>=stealth',shorten >=1pt,auto,node distance=2.3cm, scale = 1, transform shape]
  \node[state] (A) {$A_1$};
  \node[state] (B) [below left=1cm and 2cm of A] {$A_2$};
  \node[state] (C) [below right=1cm and 2cm of A] {$A_3$};
  \node[state] (D) [below left of=B] {$A_4$};
  \node[state] (E) [below right of=B] {$A_5$};
  \node[state] (F) [below left of=E] {$A_6$};
  \node[state] (G) [below right of=E] {$A_7$};

   \path[every node/.style={font=\sffamily\small,
   		fill=white,inner sep=1pt}]
     (A) edge [bend right=20] node[left=1mm] {$\frac{1}{2}.\query_{12}$} (B)
     (A) edge node[right=1mm] {$\frac{1}{2}.\query_{13}$} (C)
     (A) edge [bend right=20] node [above=1mm]{$\conn_{17}$} (G)
     (B) edge node [left=1mm]{$\frac{1}{2}.\query_{24}$} (D)
     (B) edge [bend right=20] node [left=1mm]{$\frac{1}{2}.\query_{25}$} (E)
     (B) edge [bend right=20] node [below=1mm]{$\hit_{21}$} (A)
     (E) edge node [left=1mm]{$\frac{1}{2}.\query_{56}$} (F)
     (E) edge [bend right=20] node [left=1mm]{$\frac{1}{2}.\query_{57}$} (G)
     (E) edge [bend right=20] node [right=1mm]{$\hit_{52}$} (B)
     (G) [bend right=20] edge node [right=1mm]{$\frac{2}{3}.\dwnld_{71}$} (A)
     (G) edge [bend right=20] node [right=1mm]{$\hit_{75}$} (E)
     (G) edge [loop below] node [right=1mm] {$\frac{1}{3}.\rej_7$} (G)
;
\end{tikzpicture}}
\vspace{-1mm}
    \caption{Downloading a file using Gnutella P2P network.}
    \label{fig:eg-p2p}
\end{figure}
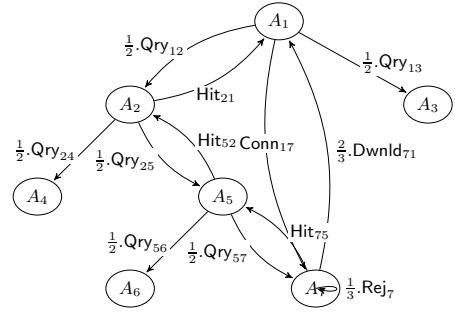

\smallskip\noindent
{\bf Example: A peer-to-peer (P2P) file sharing network.}
This case study considers a variant of a Gnutella-like P2P network for file sharing, allowing users to communicate and access files without the need for a server. The individual users in this network are referred to as peers. Gnutella protocol defines a decentralised approach making use of distributed systems, where the peers are called nodes, and the connection between peers is called an edge between the nodes, thus resulting in a graph-like structure. A peer wishing to download a file would send a query request \query \ packet to all its neighbouring nodes under a probability distribution. If those nodes don't have the required file, they pass on the query to their neighbours and so on. When the peer with the requested file is found, the query flooding stops and it sends back a query hit packet  \hit \ following the reverse path. If there are multiple query hits, the client selects one of these peers. The client thus builds a connection with the peer offering the resource and download the resource. Fig.~\ref{fig:eg-p2p} shows an example process of downloading a file using the Gnutella-like P2P network.
Let $A=\{A_i \mid i \in \{1, 2, 3, 4, 5, 6, 7\} \}$ denote a set of nodes (agents) in the network,
$\query_{ij}$ denote node $A_i$ sends a query request to node $A_j$, $\hit_{ji}$ denotes node $A_j$ sends a query hit message to $A_i$, 
$\conn_{ij}$ denotes node $A_i$ builds a connection with node $A_j$,
$\dwnld_{ji}$ denotes node $A_i$ downloads the file from node $A_j$.
Suppose $A_1$ is the agent node looking for a resource, $A_7$ is the agent node willing to offer the requested resource. 
A malicious node, say $A_4$, tries to learn some information of the situation of the downloading procedure of $A_1$.
We could ask a question ``what is the degree of the observability of $A_4$ on the procedure of $A_1$ finally downloading the requested file?'', which could be formalised in formula: 
$\P=_? \lbrack  \langle A \rangle \  \opac_4 \ \F \ ``A_1 \textit{ downloads the requested file}"\rbrack$.
Assume the observation function of $A_4$ is defined as: 
{\scriptsize \texttt{
Qry12->Q1, Qry13->, Qry24->Q2, Qry25->, Qry56->, Qry57->, Hit21->, Hit52->, Hit75->, Conn17->Con1, Dwnld71->Dwn, Rej71->Fail
}}

\smallskip
The result generated regarding to the pre-defined observation function is presented as follows:
\begin{scriptsize}
\vspace{-2mm}
\begin{Verbatim}
Result: 0.083.{
0.083:Qry12Qry25Qry57Hit75Hit52Hit21Conn17Dwnld71->Q1Con1Dwn
} (value in the initial state)           
\end{Verbatim}
\end{scriptsize}






\section{Conclusions and Future Work}
\label{sec:conc}

We have constructed a formal framework for quantitatively specifying and verifying observability properties in MASs.
Observability analysis can be used to capture information transparency and thus information leakage in MASs for information flow security concerns.
A direct application of this work is privacy loss and information leakage for security analysis in MASs.
The focus of this paper is on developing the theory and framework which provides a foundation for future work to assist operators in managing information-leakage risks while optimising performance effectiveness in collaborative multi-agent systems. Although the current implementation is a prototype for testing research ideas and demonstrating the verification framework's feasibility through small case studies, future plans include transitioning to a publicly available software tool.

In future,
we intend to integrate our observability operator into Strategic Logic~\cite{ChatterjeeHP07} and evaluate its suitability for various scenarios in information security analysis.
We will also develop novel approaches to generate policies that capture the trade-off between task completion and gathering/restricting information learned via maximising/minimising agents' observability. 
Game-theoretic methods can be integrated into the framework to automatically identify an equilibrium between information transparency guarantees and performance objectives based on the quantified results produced by this work. Such an equilibrium can be used to indicate an optimal decision for operators to coordinate behaviours in multiple domains concerning combined effectiveness and information transparency.
%
There is a wide range of applications where such abilities are necessary to balance the integrated capabilities and security risks.

\bibliography{BIB-opacMAS-full}



\end{document}